# A Demographic Attribute Guided Approach to Age Estimation

Zhicheng Cao, Kaituo Zhang, Liaojun Pang and Heng Zhao

Molecular and Neuroimaging Engineering Research Center of Ministry of Education, School of Life Science and Technology, Xidian University, Xi'an, China 710071.

## Abstract

Face-based age estimation has attracted enormous attention due to wide applications to public security surveillance, human-computer interaction, etc. With vigorous development of deep learning, age estimation based on deep neural network has become the mainstream practice. However, seeking a more suitable problem paradigm for age change characteristics, designing the corresponding loss function and designing a more effective feature extraction module still needs to be studied. What's more, change of face age is also related to demographic attributes such as ethnicity and gender, and the dynamics of different age groups is also quite different. This problem has so far not been paid enough attention to. How to use demographic attribute information to improve the performance of age estimation remains to be further explored. In light of these issues, this research makes full use of auxiliary information of face attributes and proposes a new age estimation approach with an attribute guidance module. We first design a multi-scale attention residual convolution unit (MARCU) to extract robust facial features other than simply using other standard feature modules such as VGG and ResNet. Then, after being especially treated through full connection (FC) layers, the facial demographic attributes are weight-summed by 1*1 convolutional layer and eventually merged with the age features by a global FC layer. Lastly, we propose a new error compression ranking (ECR) loss to better converge the age regression value. Experimental results on three public datasets of UTKFace, LAP2016 and Morph show that our proposed approach achieves superior performance compared to other state-of-the-art methods.

---

*Send correspondence to Dr. Heng Zhao at hengzhao@mail.xidian.edu.cn for questions.

# 1. Introduction

Face-based age estimation remains an active research topic, which is a soft biometric technology using facial feature information to determine the age of a subject. With it becoming more mature and reliable, age estimation has been widely applied to public security surveillance, human computer interface, business user management, etc.[1]

Aging can be seen as a nonlinear process since it is affected by different factors at different age period. During childhood, facial aging is mainly relevant to the change of the face shape, while in adulthood facial age changes are mainly determined by skin texture[2]. Therefore, aging is usually considered to have two characteristics: non-linearity and continuousness.

Current methods of age estimation can be divided into two categories: methods based on traditional machine learning tools and methods based on deep learning, according to the way of extracting features. Among traditional machine learning methods, the most successful hand-crafted feature method is bio-inspired feature (BIF) [3]. However, it is still difficult to design suitable features for age estimation because the mechanism of how humans perceive aging patterns at different ages is still unclear [4]. Meanwhile, since the manual feature extraction heavily relies on idea facial poses and accurate localization of facial landmarks, the robustness of traditional methods is far lower than that of the deep learning methods.

As deep learning being applied to the problem, age estimation was regarded as a task of pattern recognition. Since the value of the age is essentially an ordered number, and each age value can be regarded as a class by itself, age estimation can thus be considered as a regression problem or a classification problem. However, both models have limitations, as they tend to consider only one age characteristics. To better fit the aging pattern, the ranking model is proposed, treating aging as an ordinal regression problem.

For example, Niu et al.[4] proposed Ordinal Regression CNN (OR-CNN), which implements a sorting structure of k-1 categories with multiple output layers and each layer represents an age category, where k is the number of age categories. On this basis, Cao et al. [5] designed a consistent rank logits (CORAL) framework with theoretical guarantees for classifier consistency, which is used to solve the problem of possible inconsistency of sorting results in OR-CNN. From the perspective of the network structure, CORAL framework can be regarded as training two targets at the same time: a regression value and a series of monotonically increasing biases, and the result of the activation after the addition is used to represent the ranking label at last. There is more uncertainty in the strategy of training two targets with one loss function. So in this paper, we take a single-objective training strategy by fixing the bias, such that the network trains a regression value to represent age. Meanwhile, our label is different from the k-1 dimension of the above two methods, but is k-dimensional, which can utilize the label information of the starting age.

On the other hand, judgment of human age is highly correlated with demographic information such as gender and ethnicity, which has not been studied or paid enough attention so far. Therefore, during designing of our age estimation method, we utilize gender, ethnicity, and age group information of the original image as attribute information. It is used in the network with the concept of attribute guidance, which makes the learning process more deterministic. In brief, our contributions are:

- A more robust feature extraction module is proposed which is built on the multi-scale attentional residual convolution unit (MARCU);
- An Error Compressing Ranking Loss (ECR) is proposed to reduce the estimation error by arranging the age interval points in a more succinct way;
- An Attribute Guidance Module is proposed in order to utilize facial demographic attributes which is beneficial for age estimation;
- Experiments on different age estimation datasets show that the proposed network framework achieves state-of-the-art performance.

# 2. Related Works

In the last two decades, researchers have conducted intensive study on face image-based age estimation. The two-stage strategy consisting of feature extraction and model learning was adopted in the early days. Then, with the development of deep learning and its application to face related problems, deep learning-based and end-to-end frameworks that integrate both the stages have gradually become the mainstream practice. We hereby review the brief history of age estimation as follows.

## 2.1 Two Stage Methods

The two stages methods focus on how to extract effective discriminative features from faces, and how to effectively use the features for age estimation. Kwon et al.[2] first proposed an age estimation method by manually extracting feature points, and judging the age by proportional calculation and setting thresholds. Later, due to the development of machine learning, the form of the two-stage method gradually became a mode of manually extracting features and using machine learning methods for processing and discrimination. AAM[6] is the earliest method to extract shape and appearance features of face images. Then BIF[3], as the most successful age feature, was widely used for age estimation. In the second stage, classification and regression are usually used to estimate facial attributes, the former includes k-nearest neighbors (KNN), multilayer perceptron (MLP) and support vector machine (SVM), the latter includes quadratic regression, and support vector regression (SVR).

There are many limitations in the manual feature extraction method at the first stage. For example, when facing new tasks, it is often necessary to redesign features, which requires a lot of professional knowledge and related attempts. In terms of age estimation, it is often difficult to design well-targeted manual features since the mechanism of how human perceive aging is still unclear, while the effectiveness of the second stage depends on high-precision feature extraction, which makes the robustness of the two-stage method far less than that of the deep learning model under big data, thus the two-stage method is gradually replaced by the end-to-end deep learning method.

## 2.2 Deep Learning Methods

By performing both feature extraction and age judgment in an end-to-end fashion, deep learning has achieved state-of-the-art results on multiple visual recognition tasks. In the field of age estimation, deep learning methods can be divided into four categories: metric regression [7], multi-class classification[8], deep label distribution learning [9, 10], and ranking methods[4, 5, 11].

Metric regression treats age estimation as a real-valued regression problem. The training process usually minimizes the Euclidean distance between the estimated values and ground truth values.

Multi-class classification adopts a general image classification framework, takes the output of the neuron as the probability of the corresponding age category, and maximizes the probability of the true category as the goal during the training process. In the inference stage, Rothe et al.[8] empirically show that the expected value of the output probability achieves better performance than predicting the class with the largest probability. However, both metric regression and multi-class classification often lead to unstable training[9].

Deep label distribution learning transforms a single value into a series of label distributions, expressing the desired value in the form of a weighted sum of distributions and probabilities. Gao et al.[9] learns the label distribution in an end-to-end manner and is used to solve problems such as semantic segmentation, age estimation, object detection, etc. On this basis, Gao et al.[10] pointed out that in paper[9], the training objective and the evaluation index are inconsistent, and the result is sub-optimal. Therefore, after obtaining the result of the learned label distribution, through weighted training again, they obtain results consistent with the evaluation metrics.

The ranking methods treats age estimation as an ordered regression problem, which is then transformed into a series of binary classification problems, representing the ordinal regression result as an aggregation of the results of multiple binary classifiers. Niu et al.[4] proposed a multi-output CNN called OR-CNN by integrating multiple binary classifiers into the CNN. Later Chen et al.[11] proposed a modification of OR-CNN to train a series of binary classification CNNs, called Ranking-CNN, to obtain better results. Cao et al.[5] designed a monotonic sorting network structure based on Niu, aiming at the possible inconsistency of its classifier results. The results are closer to the original intention of the sorting design idea, and achieved the most advanced results.

# 3. Proposed Approach

This section describes our proposed approach to age estimation from the perspectives of loss function, network framework and the attribute guidance module.

## 3.1 Loss Function

Aiming to improve the problem of training multiple targets in CORAL[4], we propose an Error Compressing Ranking (ECR) loss function。In order to achieve the consistency of the ranking results, the CORAL network actually trains two targets：The weight parameter of the front layer shared by all classifiers, and the monotonically non-increasing biases added by each classifier individually。Both serve one result, but in different ways. In general, networks training multi-objective has more uncertainty in the training process. CORAL uses different random seeds in the experimental stage to ensure that the network can converge effectively. When the monotonically increasing offset is fixed here, the offset can exist away from the network, so that the network can achieve the goal of training a single regression value.

To keep the expression of the regression value as simple as possible, we consider setting it as the predicted age output, and on this basis, further considering the design of offset. There has been a problem in ranking model with sorting labels that is for *K* age categories, the dimension of the sorted labels is *K-1*. The estimated age is obtained by adding the number of positive classifiers to the starting age, where the positive result of classifier means the output is greater than its represented age. This is actually the default for all images to be at least older than the starting age, which is logical in the original intention of the sorting algorithm design, but on the other hand, it can also be understood that the information of the starting label is not used, and the age less than the starting tag cannot be judged. Therefore, for *K* age categories, we designed *K* dimension segment point, which we call the age interval points, and the corresponding real order label dimension is also *K*.

In order to ensure the network effectively uses the sorting label to output the age regression value, the age interval points start from 0.5 to the left as the starting age value, and increases by 1 in sequence until the number of total intervals is the same as the total classes of age.

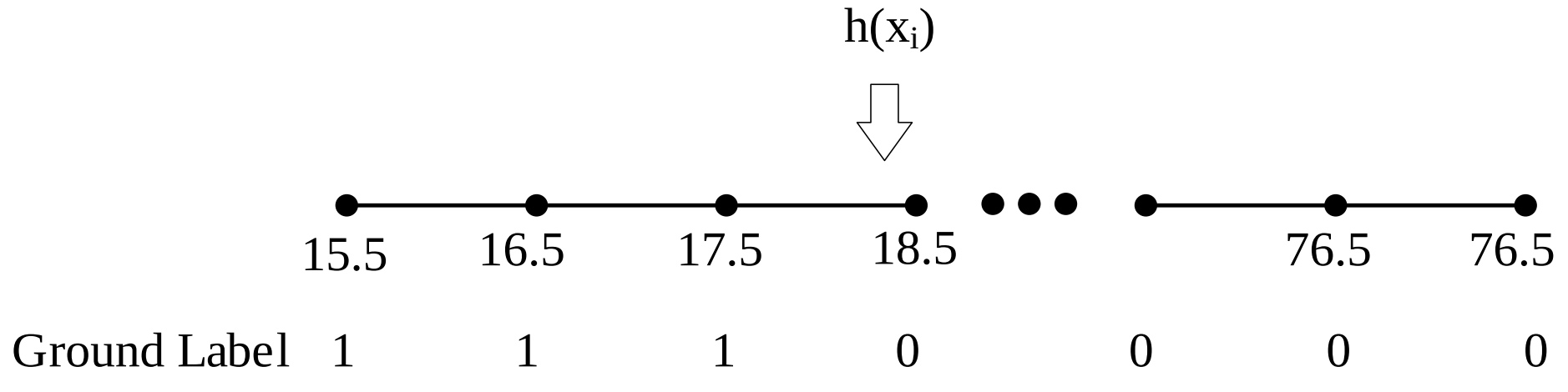

Feature 1. Age interval points labeling with example of the Morph dataset.

A specific example for the idea of age interval points on the Morph dataset are shown in Figure 1, where the age distribution of the Morph dataset is from 16 to 77 years old. Therefore, the age interval points is set from 15.5 to 76.5 with an interval of 1, which ensures that the number of intervals is the same as that of the age classes.

The advantage of this setting is that the adjacent 1 and 0 in the sorted labels correspond to the ground value ±0.5, thus the effective convergence of cross-entropy can limit $h(x_i)$ to the interval between 1 and 0 labels, that is, the ground value ±0.5. At the same time, the training method of the regression value output allows the age to appear at any position on the coordinate axis. Compared with the order label of the dimension, this design can use the information of the starting age label. Furthermore, the decimal output makes the calculation of the error between the result and the ground value more precise.

By designing the age interval points at both ends of the ground age value, we combine the ranking label to make the network output a regression value to represent the age, which we call error-compression ranking loss (ECR) loss. The specific calculation equation is as follows:

$$L_{ecr} = -\sum_{i=1}^{N}\sum_{k=1}^{K}[\log(\sigma(h(x_i)-b_k))y_i^{(k)} + \log(1-\sigma(h(x_i)-b_k))(1-y_i^{(k)})] \quad (1)$$

Where $x_i$ is the input image, $h(x_i)$ is the regression value output by the network model, $b_k$ is the K age interval points, K is the number of age categories, and $y_i$ is the ground ranking label.

## 3.2 Network Framework

During the design of the basic convolution block for facial feature extraction, based on the ResNet[12] residual block, we add a multi-scale feature extraction module[13] and a channel attention mechanism[14] to obtain a Multi-scale Attentional Residual Convolution Unit. Its structure is shown in Figure 2:

In the design of the multi-scale feature extraction module, we use parallel convolution kernels of different sizes to perform convolution operations with different channel numbers on the front-layer feature maps. The specific details are to perform $1\times1$ convolution 1/4 of the number of output channels, $3\times3$ convolution of 1/2 of the number of output channels, and $5\times5$ convolution of 1/4 of the number of output channels for the input feature map in turn, and then splice the channels to obtain Output feature map.

After that, we calculate the channel weights on the feature map to realize the channel attention mechanism. In this step, we adopt the idea of ECA structure[15], that is, after pooling the feature map globally, a weighted fully connected layer is obtained, then one-dimensional convolution is performed on the fully connected layer, and finally activate operation obtains a set of reassigned value, to represent feature map channel weights. Its structure is shown in Figure 3.

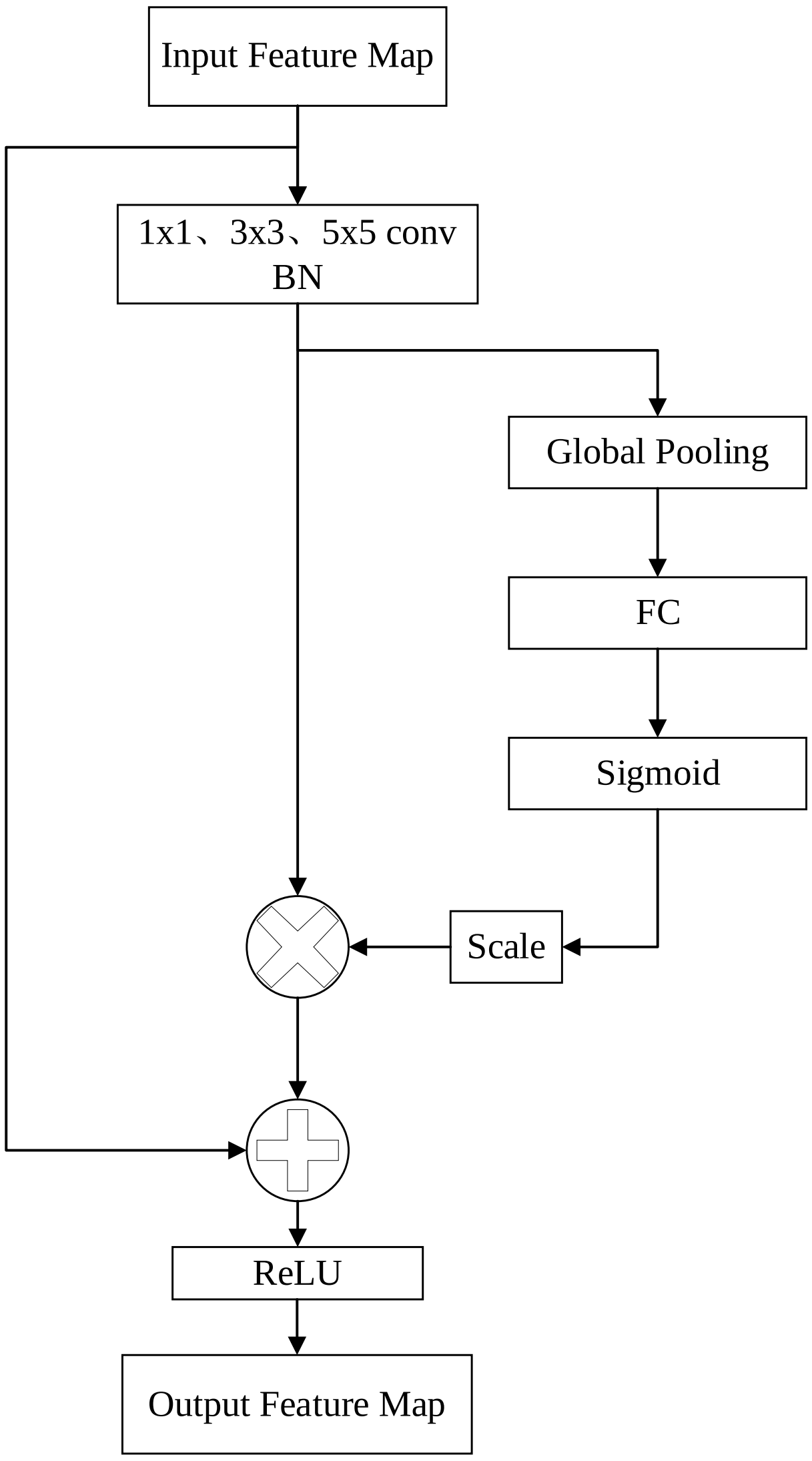

Feature 2. The structure of the proposed Multi-scale Attentional Residual Convolution Unit.

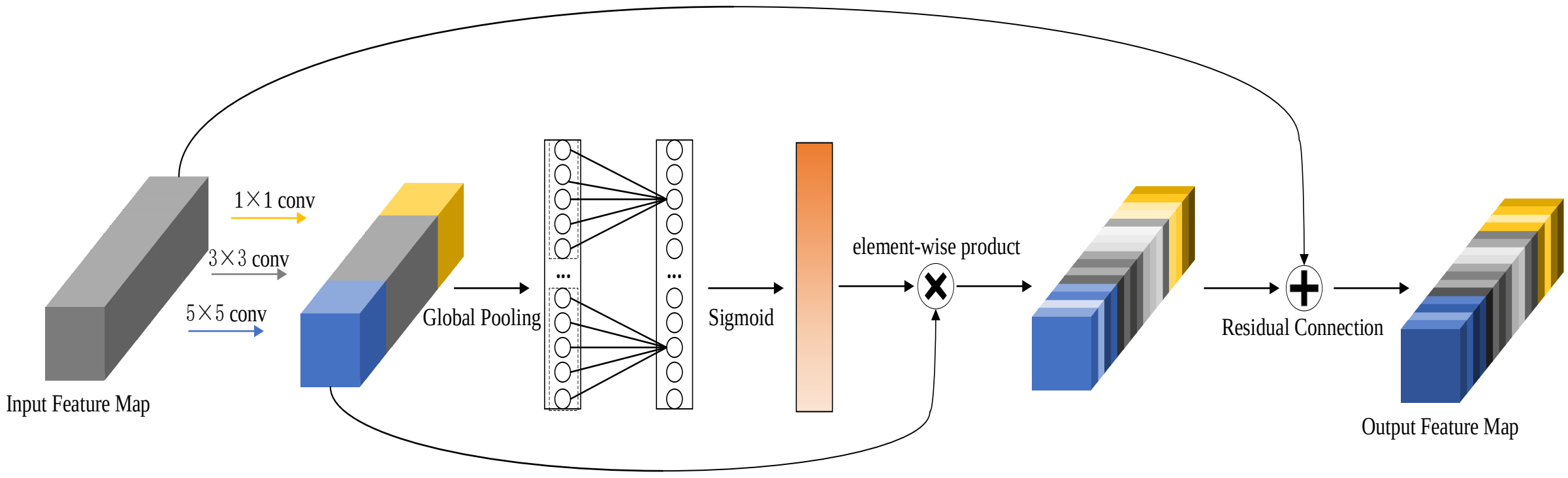

Figure 3. Channel attention module explained.

As seen from Figure 3, after multi-scale convolution is performed on the input feature map, the channel attention weight is calculated. At this time, instead of the calculation method of the fully connected layer, we use the one-dimensional convolution method to utilize the direct information between adjacent channels. is multiplied element by element with the feature map after obtaining the channel attention weight, and finally the residual connection is performed with the input feature map to obtain the output feature map. The channel attention weights are multiplied element by element with the feature map, and finally the residual connection is performed with the input feature map to obtain the output feature map. When performing one-dimensional convolution, we set convolution kernels of different sizes according to the number of specific feature map channels. Specifically, when the number of channels is small, we set a convolution kernel of $1\times3$ to perform one-dimensional convolution, and when the number of channels is large, convolution kernel is set of $1\times5$.

Between different MARCU connections, the feature map adopts down sampling operation to keep the input and output feature map size consistent. In order to ensure the comparability of the feature extraction structure, we use [6, 8, 12, 6] blocks to stack MARCU to ensure the same number of convolutional layers as ResNet34. The specific parameters are shown in the following table:

Table 1. Parameters of the proposed feature extraction module of our age estimation network.

| Layer | Name | Function | Kernel | Output feature map size |
|---|---|---|---|---|
| **1** | Conv1 | convolution | 7×7,64, stride 2 | 112×112 |
| **2** | Maxpool1 | pooling | 3×3 stride 2 | 56×56 |
| **3~8** | Conv2×6 | convolution | 1×1,16<br>3×3,32<br>5×5,16<br>1×3 FC | 56×56 |
| **9~16** | Conv3×8 | convolution | 1×1,32<br>3×3,64<br>5×5,32<br>1×3 FC | 28×28 |
| **17~28** | Conv4×12 | convolution | 1×1,64<br>3×3,128<br>5×5,64<br>1×5 FC | 14×14 |
| **29~34** | Conv5×6 | convolution | 1×1,128<br>3×3,256<br>5×5,128<br>1×5 FC | 7×7 |

## 3.3 Attribute Guidance Module

After the design of the loss function and the feature extraction module, we further consider another issue of age estimation, that is, judgement of human age via the face is affected by other demographic attributes, such as gender and ethnicity. Therefore, we propose to utilize other attribute information of face when designing the network, and let the attribute information guide the estimation result. By calculating the loss between feature information and attributes, the expression of features is limited, so as to strengthen the judgment of results. Such network is more deterministic during training.

Inspired by the literature[16, 17], we use attribute information to limit the final feature fully connected layer when using the network for age discrimination, and then assists the judgment of the age. The structure is designed as follows: the fully connected layer of the network after feature extraction is expanded and input into branches of different dimensions, and the calculation of different attribute information is performed separately. Then after splicing it, a new same-dimensional fully connected layer is connected to obtain the attribute features for the final calculation. When designing the total loss, all attribute losses are added together.

When using attribute features, this paper considers further feature extraction for the fully connected layer. The design here is based on two considerations. On the one hand, due to the phenomenon that age is highly correlated with other face attributes, attribute features need to be used in the network. Attribute features are generally based on a fully connected layer that calculates the loss with the corresponding attribute label. Therefore, when further feature extraction is performed on the fully connected layer, a new fully connected layer can be added on the basis of the original structure; on the other hand, the fully connected layer of attribute features is the information at the relative result level in the sense. When adding a fully connected layer, if there are too many layers, the fully connected layer of features will lose its meaning. Therefore, when designing a new fully connected layer for feature extraction, only one fully connected layer is considered.

Similar to the idea of SE-NET in literature[15], the connection method of the fully connected layer will make the local information utilization of the original attribute fully connected layer become the global information utilization, which is not beneficial to the enhancement of information. Therefore, after obtaining the fully connected layer after the splicing of the attribute fully connected layer, a one-dimensional convolution method is used to obtain the second attribute fully connected layer and the second attribute fully connected layer, and the final structure of our proposed method (referred as Attribute-Guided AgeNet or simply as AG-AgeNet) is obtained. Its structure is shown in Figure 4:

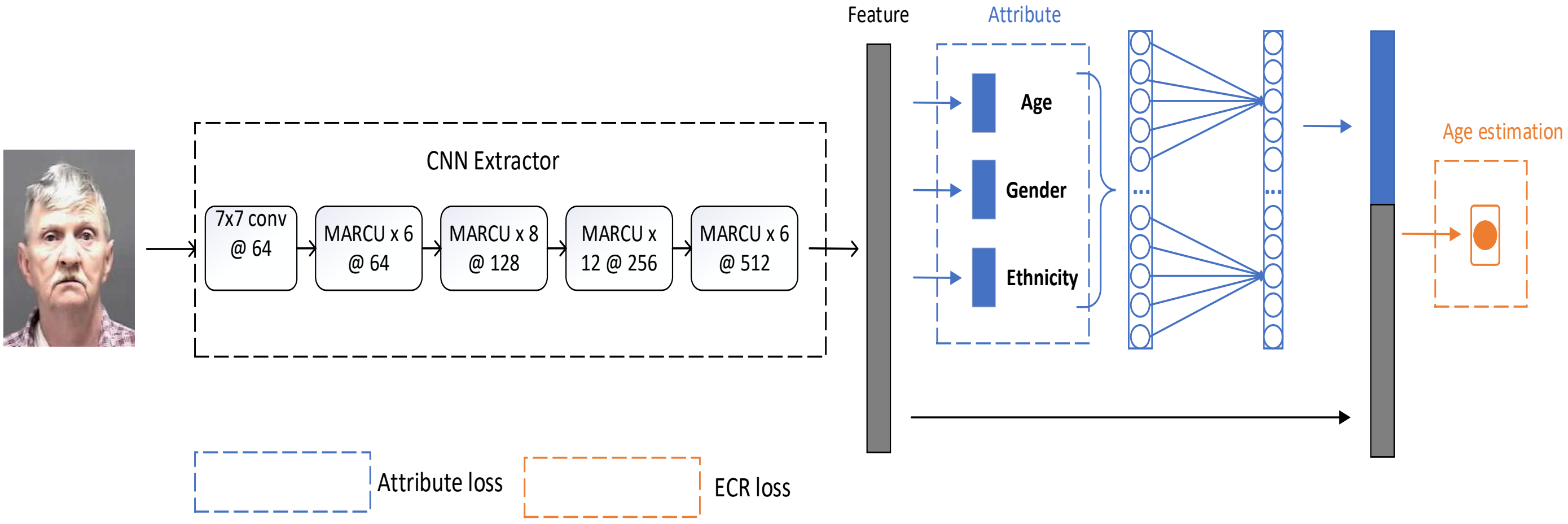

Figure 4. Overall structure of the proposed Attribute-Guided AgeNet.

AG-AgeNet comprises three fully connected layers with different attributes and a global fully connected layers, where the classification results of the gender attribute layer, the age group attribute layer, and the ethnicity attribute layer are calculated by cross-entropy respectively with the ground label of the corresponding information. The network splices the three attribute fully connected layers, and calculates in one-dimensional convolution to obtain the second attribute layer of the same dimension. Then, the second attribute fully connected layer is spliced with the global fully connected layer to obtain the final fully connected layer. Finally, the calculation of age estimates is performed using the ECR loss.

The total loss function is the sum of attribute loss and age estimation loss, whose expression is as follows:

$$L_{total} = L_{ecr} + L_{attr} \tag{2}$$

$$L_{attr} = -\alpha \sum_{i=1}^{N} [\log(a(x_i)a_i)] - \beta \sum_{i=1}^{N} [\log(g(x_i)g_i)] - \gamma \sum_{i=1}^{N} [\log(e(x_i)e_i)] \tag{3}$$

where $a(x_i)$ is the calculated age group classification, $a_i$ is the image age group label, $g(x_i)$ is the calculated gender classification, $g_i$ is the image gender label, $e(x_i)$ is the calculated ethnicity classification, and $e_i$ is the image ethnicity label. $\alpha$, $\beta$ and $\gamma$ are the attribute coefficients, which can be adjusted according to specific data bias. Throughout the experiments of this paper, $\alpha$, $\beta$ and $\gamma$ are all set to 1.

# Experiments and Analysis

## 4.1 Datasets

The first dataset, Morph II, is one of the most well-known and publicly available age estimation datasets, which contains 55,608 images of 13,000 individuals. Since the acquisition conditions such as lighting and pose of this dataset are relatively uniform, the image style is consistent and the image quality is good. The ages in the dataset range from 16 to 77 years old, and the labels include the age and other demographic information of the subjects such as gender and ethnicity.

The second dataset, UTKFace, is another widely-used age dataset with a large age span, containing more than 20,000 face images and ranging from 0 to 116 years. The labels of the dataset contain demographic information of the age, gender, and ethnicity of the people. The dataset has a wide range in terms of pose, facial expression, illumination, resolution, etc. It has also been used for many other face related tasks according to the specific need.

The third dataset of LAP2016 is an apparent age dataset collected by the ChaLearn Looking at People team for the 2016 Face Apparent Age Estimation Challenge. The age span is from 0 to 100 years old, and the total number of images exceeds 8,000. Different from other age dataset, apparent age refers to the age judgment of the image by the human eye, i.e., "how old it looks". The label of each face is estimated by several individuals at the same time, such that the image label provides a variance in addition to the mean of age.

## 4.2 Implementation Details

In terms of environment configuration, the operating system used in this experiment is window 10, the CPU is Xeon W-2133, the GPU is RTX2080 Ti, and the deep learning environment is Anaconda3 with Pycharm 2019.3.4.

During processing of the datasets, firstly, the Morph dataset and the UTKFace dataset are randomly divided into subsets of training, validation, and test with a ratio of 8:1:1. For the UTKFace dataset, only 1 to 100 age labels are used. The LAP2016 dataset is divided according to the official protocol of training, validation, and test. Due to the age distribution issue, only images with age labels from 3 to 80 are used for experiments.

After getting the training set, validation set and test set, the images are preprocessed. Operations such as face detection, cropping, and scaling are included. For face detection and cropping, we use the detection function and cropping function in the *dlib* public library in the field of image processing. After detecting the face image, the image with the detected face number of 1 is retained, and the image is cropped. Then the cropped image is scaled, and the size of the image set is uniformly scaled to 256×256 resolution in the experiment.

During training, the images in the training set are randomly cropped to a size of 224×224 as the network input to prevent overfitting. A total of 200 iterations were performed for training. The Adam algorithm was used as the optimizer, the learning rate was fixed at 0.0005, and the batch size was set as 64.

The strategy of retaining model is to input the images in the training set into the network in batch size, and after forward calculation, the learned parameters are obtained by reverse optimization of the loss function. After one iteration of the training set, the results of the model on the validation set are calculated. The judgment at this time is based on the MAE between the estimated value and the ground age label of the validation set. When the newly calculated MAE is less than the current recorded value, update the model parameters, record the MAE, and proceed to the next iteration until the total number of iterations is reached. The last reserved model parameters are used for the test set calculation results for performance evaluation.

## 4.3 Evaluation Metrics

MAE is the most commonly used and the most basic evaluation performance metricsin the field of age estimation. It refers to estimating the age of the test image set, subtracting the result from the true value to obtain the absolute value, and finally averaging the absolute error. Its calculation formula is as follows:

$$MAE = \frac{1}{N}\sum_{k=1}^{N} | \hat{y}_i - y_i | \tag{4}$$

where $\hat{y}_i$ represents the model estimation result of the $i$th sample, $y_i$ represents the ground label of the sample, and $N$ represents the number of images in the test dataset. It is generally believed that the smaller $MAE$ means better performance of the model. Therefore, reducing $MAE$ of the model estimation results on the test data set is often used as the goal of model training.

## 4.4 Comprehensive Experiments

### 4.4.1 Experiment on the ECR Loss

To illustrate the effectiveness of the proposed ECR loss function, we conduct ablation experiments using the same feature extraction network but with different age expressions and corresponding loss functions.

For the setting of the methods of comparison, due to the good feature extraction performance of ResNet34, we uniformly use ResNet34 as the feature extraction module if the age expression that does not limit the feature extraction network structure in the original text. For DLDL-v2, the ThinAgeNet and TinyAgeNet models of the original paper are used.

The age estimation results on the test set are analyzed in terms of the MAE metrics which are shown in Table 2. From the table, we can see that the method using our proposed ECR loss performs very well in terms of the MAE metrics. Specifically, on both the UTKFace dataset and LAP2016 dataset, it was the proposed method of ECR-CNN that achieved the best results, which are 3.61 and 6.29, respectively. On the Morph dataset, ECR-CNN achieved a suboptimal result of 2.42 and CORAL-CNN achieved the best result of 2.39. To summarize, the proposed method of ECR-CNN achieves better or comparable results to other ranking methods on all three datasets, suggesting the superiority of our proposed loss function.

Table 2. Comparison of MAE between our age estimation method with ECR loss and other methods in the literature.

| Method | Year | Feature Extraction Module | Dataset | | |
|---|---|---|---|---|---|
| | | | Morph | UTKFace | LAP2016 |
| CE-CNN | - | ResNet34 | 2.73 | 6.14 | 8.52 |
| MAE-CNN | - | ResNet34 | 2.48 | 3.67 | 6.97 |
| OR-CNN | 2016 | ResNet34 | 2.46 | 5.37 | 6.36 |
| CORAL-CNN | 2020 | ResNet34 | **2.39** | 5.17 | 7.08 |
| DLDL-v2 | 2020 | TinyAgeNet | 2.76 | 4.36 | 6.72 |
| DLDL-v2 | 2020 | ThinAgeNet | 2.46 | 4.03 | 6.53 |
| **ECR-CNN (Proposed)** | 2021 | ResNet34 | 2.42 | **3.61** | **6.29** |

It is also worth mentioning that the network structure of ECR-CNN and MAE-CNN used in this paper use the same network structure. It is observed that the MAE results of ECR-CNN on the three datasets are all better than the results of MAE-CNN. This indicates that for a nonlinear regression problem, the proposed ECR loss can better represent the nonlinearity in the subject change process than the common MAE loss.

### .4.2 Experiment on the MARCU Module

When describing the performance of the designed feature extraction network, we first determine that age estimation is the goal, and the calculation of the loss function uses the ECR loss uniformly, which shows that the proposed feature extraction network is beneficial to the designed loss function. In the setting of the comparative experiment, we select some classic network architectures in the field of deep learning, including VGG19, DenseNet, ResNet50 and ResNet34. The specific values are shown in the Table 3.

Table 3. MAE of age estimation using different feature extraction modules.

| Dataset | VGG19 | DenseNet | ResNet50 | ResNet34 | **MARCU (Proposed)** |
|---|---|---|---|---|---|
| Morph | 3.25 | 3.73 | 2.54 | 2.42 | **2.38** |
| UTKFace | 4.57 | 8.67 | 3.80 | 3.61 | **3.55** |
| LAP2016 | 6.42 | 7.73 | 6.53 | 6.29 | **6.04** |

VGG19, ResNet50 and ResNet34 are implemented in the original structure, except for replacing the last fully connected layer to single-valued output for age estimation. DenseNet is configured as 3 Dense blocks, each block has 4 layers of convolutional layers, and each layer outputs 4 feature map.

It can be seen that the performance of the proposed MARCU is improved compared to ResNet34 on the three datasets while maintaining the same age expression and loss function calculation. And it achieves the best results in different comparison structures, specifically 2.38 in Morph, 3.55 in UTKFace, and 6.04 in LAP2016. This shows that for the age estimation task, when ECR loss is used for regression value calculation, the multi-scale feature extraction and channel attention mechanism adopted by the proposed MARCU enables the MARCU network to effectively complete the goal of feature extraction, improving the overall network performance.

### .4.3 Ablation Experiment for Attribute Guidance

After obtaining the results of MARCU, we conducted ablation experiment for the attribute guidance module proposed in this paper.

In the division of age groups, we add adolescent age groups based on the World Health Organization's age classification standards for experimental research and analysis. The specific classification criteria are: young people under 18 years old, young people between 18 and 44 years old, middle-aged people between 45 and 59 years old, old people between 60 and 74 years old, and senior citizens between 75 and 89 years old. and above are the long-lived elderly. The number of face images in different age groups in each dataset used in the experiment is shown in Table 4.2. Combining datasets with different age distributions, the label settings for the experiments are given below.

Table 4. Demographic information of ethnicity and gender in each dataset.

| Dataset | Ethnicity | Gender |
|---|---|---|
| LAP2016 | White | Men、Women |
| Morph | White、Black、Latino、Asian | Men、Women |
| UTKFace | White、Black、Asian、Indian | Men、Women |

Table 5. The specific distribution of faces corresponding to different age groups in each dataset.

| Dataset | Age Group | | | | | | | | | | | |
|---|---|---|---|---|---|---|---|---|---|---|---|---|
| | 1-17 | | 18-44 | | 45-59 | | 60-74 | | 75-89 | | ≥90 | |
| LAP2016 | 3224 | 6.4% | 39370 | 77.6% | 7841 | 15.4% | 298 | 0.6% | 5 | 0.01% | 0 | 0% |
| Morph | 3914 | 17.7% | 12508 | 56.7% | 3173 | 14.4% | 1581 | 7.2% | 755 | 3.4% | 126 | 0.6% |
| UTKFace | 254 | 6.4% | 3048 | 77.4% | 529 | 13.4% | 103 | 2.6% | 8 | 0.2% | 0 | 0% |

In the Morph dataset, the tags contain the age, gender, and ethnicity information of the subjects. Therefore, after the age is set as the ranking label, the gender and ethnicity are respectively set as different attribute information for the realization and analysis of the subsequent network results. In the setting of age groups, since the age span of this dataset is 16 to 77 years old, according to the distribution of the number of images in different age groups, the ages are divided into 3 groups, 16 to 44, 45 to 59, and 60 to 77. The UTKFace dataset also contains the age, gender and ethnicity information of the subjects, so the attribute information settings of gender and ethnicity are consistent with the Morph dataset. In the age group setting, we only use the age from 1 to 100 years old for the experiment. Based on the distribution of the number of images in different age groups, the age is divided into 6 categories: less than 18, 18 to 44, 45 to 59, 60 to 74, 75 to 89, 90 to 100. The LAP dataset only contains the age information of the subjects, so all other attribute information is manually annotated. The age interval of the LAP2016 dataset used for the experiments in this paper is from 3 to 80. We divided the age into 5 categories based on age less than 18, 18 to 44, 45 to 59, 60 to 74, 75 to 80. Since the ethnicity of the faces in this dataset is basically white, no ethnic information is marked. In the labeling of gender information, this paper uses the gender model trained by DeepFace[18] to judge the overall image, and uses the result as an attribute label for subsequent experiments.

The MAE results of the overall structure on the test set of three different datasets are shown in the Table 6. It can be seen that on the MAE of the three datasets, the structure of AG-AgeNet finally reached the optimum on the three datasets, which were 2.32 for the Morph dataset, 3.51 for the UTKFace dataset, and the LAP2016 dataset. of 5.84. This shows that the addition of the attribute guidance module can effectively improve the performance of the overall network. When using the network for age estimation, the fully-connected layers of age, ethnicity, and gender attributes are combined to make the training more deterministic and attribute constraints, which can strengthen the judgment of the results.

Table 6. Module ablation experiments for the attribute guidance.

| Dataset | AG-AgeNet without the AG module | **AG-AgeNet (Proposed)** |
|---|---|---|
| Morph | 2.38 | **2.32** |
| UTKFace | 3.55 | **3.51** |
| LAP2016 | 6.04 | **5.84** |

# 5. Summary

This research studies the problem of face-based age estimation and introduces a new deep learning-based approach. We propose to improve the age estimation performance by designing a more robust feature extraction module, involving an error compression loss, and considering the guidance of other demographic attributes. For the feature extraction module, we design a multi-scale residual convolution unit with attention mechanism (MARCU). For the error compression loss, we introduce a new way of arranging the age interval points which successfully compresses the ranking loss. Finally, we design an attribute guidance module to utilize useful demographic attribute information. Experiments on different datasets justify the superiority of our new approach of age estimation over other state-of-the-art methods in the literature.